\documentclass[5p]{elsarticle}

\usepackage[utf8]{inputenc} 
\usepackage[T1]{fontenc}    

\usepackage{hyperref}       
\usepackage{url}            
\usepackage{booktabs}       
\usepackage{nicefrac}       
\usepackage[english]{babel}
\usepackage{amsmath, amssymb, amsfonts, amsthm}
\usepackage{graphicx}
\usepackage{color}
\usepackage[capitalize]{cleveref}

\makeatletter
\def\ps@pprintTitle{%
  \let\@oddhead\@empty
  \let\@evenhead\@empty
  \let\@oddfoot\@empty
  \let\@evenfoot\@oddfoot
}
\makeatother

\hypersetup{
    colorlinks,
}

\graphicspath{{./plots/}}

\DeclareMathOperator{\real}{\operatorname{Re}}
\DeclareMathOperator{\imag}{\operatorname{Im}}
\DeclareMathOperator{\dft}{\operatorname{DFT}}
\DeclareMathOperator{\idft}{\operatorname{DFT}^{-1}}

\begin{document}
\begin{frontmatter}
\title{Explainable AI for Time Series via Virtual Inspection Layers}

\author[hhi]{Johanna Vielhaben}
\ead{johanna.vielhaben@hhi.fraunhofer.de}
\author[hhi]{Sebastian Lapuschkin}
\ead{sebastian.lapuschkin@hhi.fraunhofer.de}
\author[fu,tu,bifold]{Gr{\'e}goire Montavon}
\ead{gregoire.montavon@fu-berlin.de}
\author[hhi,tu,bifold]{Wojciech Samek\corref{cor1}}
\ead{wojciech.samek@hhi.fraunhofer.de}
\address[hhi]{Department of Artificial Intelligence, Fraunhofer Heinrich Hertz Institute, 10587 Berlin, Germany}
\address[tu]{Department of Electrical Engineering and Computer Science, Technische Universit\"at Berlin, 10587 Berlin, Germany}
\address[fu]{Department of Mathematics and Computer Science, Freie Universit\"at Berlin, 14195 Berlin, Germany}
\address[bifold]{BIFOLD -- Berlin Institute for the Foundations of Learning and Data, Berlin, Germany}
\cortext[cor1]{Corresponding author}

\begin{abstract}
The field of eXplainable Artificial Intelligence (XAI) has greatly advanced in recent years, but progress has mainly been made in computer vision and natural language processing. For time series, where the input is often not interpretable, only limited research on XAI is available. In this work, we put forward a \textit{virtual inspection layer}, that transforms the time series to an interpretable representation and allows to propagate relevance attributions to this representation via local XAI methods like layer-wise relevance propagation (LRP). In this way, we extend the applicability of a family of XAI methods to domains (e.g. speech) where the input is only interpretable after a transformation. Here, we focus on the Fourier transformation which is prominently applied in the interpretation of time series and LRP and refer to our method as \textit{DFT-LRP}.
We demonstrate the usefulness of \textit{DFT-LRP} in various time series classification settings like audio and electronic health records. We showcase how DFT-LRP reveals differences in the classification strategies of models trained in different domains (e.g., time vs.\ frequency domain) or helps to discover how models act on spurious correlations in the data.
\end{abstract}

\end{frontmatter}

\section{Introduction}
The field of XAI has produced numerous methods that shed light on the reasoning processes of black box machine learning models, in particular deep neural networks. 
\textit{Local} XAI methods quantify the contribution of each input feature toward the model output on a per-sample basis.
Prominent examples like LRP \cite{BachPLOS15}, Integrated Gradients \cite{sund_ig_2017}, LIME \cite{ribeiro_lime_2016} or SHAP \cite{lundberg_shap_2017} provide valuable insights into the intricate decision function of a neural network.
The feature-wise relevance scores they produce are usually presented as a heatmap overlaying the sample \cite{jeyakumar_how_2020}, such that they guide the eye to the important parts of the sample. In this way, it is the human user who makes the actual interpretation, e.g. "The model focuses on the dog's ears.".
These explanations work well for images or text, where XAI methods can rely on the visual interpretability of feature relevance scores.
Here, we see a reason why most XAI methods were developed and tested for applications in computer vision or natural language processing domains.
The implicit requirement of feature interpretability is challenged for time series, where single or collective time points are often not meaningful for humans \cite{samek_XAI_2021}. To exemplify,
consider the simple case of a model that classifies the frequency of a single sinusoid. Here, it is not important which minima or maxima the XAI method highlights, but how far the highlighted features are apart. In the more realistic case of a superposition of multiple sinusoids, it is impossible for the human user to derive the classification strategy from the heatmap. We see this as a reason, why only limited XAI research is available for time series \cite{rojat_xaisurvey_2021}. 

\begin{figure*}
	\center
    \includegraphics[width=0.8\linewidth]{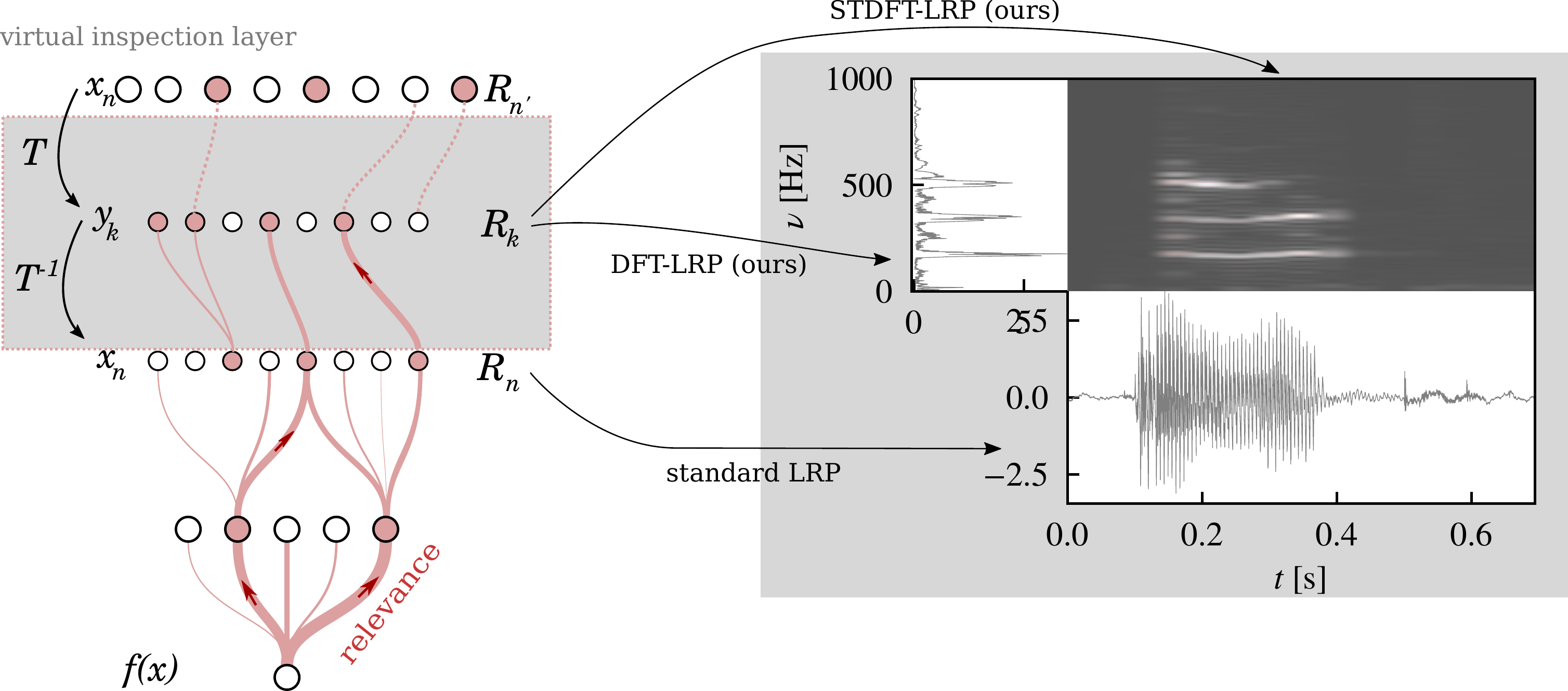}
	\caption{Schematic overview of virtual inspection layers and (ST)DFT-LRP. \textbf{Left}: A virtual inspection layer is inserted before the original input layer, performing a transformation $\mathcal{T}$ of the original input data $x_n$ to an interpretable representation $y_k$ and back. Relevance is propagated from the output $f(x)$ to the original input $x_n$ via LRP to arrive at relevance scores $R_n$. These are then propagated further through $\mathcal{T}^{-1}$ for relevance scores $R_k$ on the interpretable representation $y_k$. \textbf{Right}: Explanations for sex classifier operating on raw waveforms of voice recordings. With standard LRP, only relevance in the time domain (lower panel) is accessible, which is distributed rather uniformly over the part of the signal with large amplitude, making it impossible to derive a classification strategy. For time series, DFT or STDFT as choices for $\mathcal{T}$ lead to DFT-LRP and STDFT-LRP and relevances in frequency (left panel) or time-frequency (center panel) domain. Only here it becomes apparent that the model is focusing on a fundamental frequency between 180 and 200 Hz and subsequent harmonics which is typical for female voices.   }
    \label{fig:scheme}
\end{figure*}

In this work, we introduce the idea of improving the interpretability of explanations for time series, by propagating them to an interpretable representation via a \textit{virtual inspection layer}.
A natural choice for an interpretable representation of time series is in the frequency or time-frequency domain. These domains are connected to the time domain via linear invertible transformations, namely the Discrete Fourier Transformation (DFT) and the Short Time Fourier Transformation (STDFT). We leverage this to propagate relevance scores for models trained in the time domain into the frequency or time-frequency domain without model re-training and without causing any change to the decision function. See \Cref{fig:scheme} for an illustration of an audio signal and model relevances in all three domains. 
This idea generalizes to any other invertible linear transformation of the data ${x}$ or a representation in latent feature space that renders it interpretable.

Technically, we attach two linear layers to the input: one that transforms the data from the original representation to the interpretable representation, and one that transforms it back (unmodified) to the original data format. Then, any local XAI method can be used to quantify relevance in the {\it new} interpretable input domain. Here, backpropagation-based methods like LRP have the advantage, that one needs to propagate the relevance scores only one layer further, i.e., just the original relevance scores are required instead of the entire model. For this reason and because of its successful application in a wide range of domains \cite{samek_XAI_2021}, we will focus on LRP in this work and refer to our method as \textit{DFT-LRP}. However, we would like to stress that our idea of the virtual inspection layer can be combined with any other local XAI technique.

We see applications of our method in particular for acoustic or sensory data where raw time series features, i.e., single time points, are particularly hard to interpret.
First, our approach can be employed to render the explanations for an existing model trained in the time domain more interpretable, without the need to retrain a model in the other domain. In other words, our approach allows interpreting a given model in (the original) time domain as well as in (the virtually constructed) frequency or time-frequency domain, practically without any additional overhead.
Second, we can compare the strategies of models trained in different domains on the same representation of the data. In particular in audio classification, finding the best input data representation, i.e. raw waveforms vs.\ spectrograms with different filters, is an important research question \cite{gupta_esc, dl_audio}. 
Here, our approachiele provides a well-informed basis for the selection of the final model, based on the model strategies (and their alignment with prior knowledge) and not only based on accuracy.

Our contributions are the following:
\begin{itemize}
    \item We propose a new form of explanation for models trained on time series data, that highlights relevant time steps as well as frequencies.
    \item We present a closed-form expression for relevance propagation through DFT and STDFT.
    \item We extend pixel-flipping-based evaluations to allow for a comparison of explanations in different formats (time, frequency, or time-frequency).
    \item We showcase how DFT-LRP gives insights on ML model strategies of audio and ECG classifiers in frequency domain, and how it can reveal Clever Hans strategies \cite{cleverhans}.
\end{itemize}
In summary, we put forward a virtual inspection layer that allows for explanations of ML models whose inputs are not directly interpretable. Our method does not require any model retraining or approximation.
When used in combination with backpropagation-based methods such as LRP our method simply requires propagating one layer further. Our method is however also applicable alongside a broad family of local XAI methods, thereby widening the general applicability of XAI to time series models.

\section{Related Work}
Prominent applications of deep learning for time series modeling are in the domain of audio processing\cite{dl_audio, gupta_esc, covid_speech}, electronic health records like ECG or EEG \cite{sturm_eeg_2016, ecg_spectroram_classifier, eeg_tsc_21} or forecasting in domains like finance \cite{financial_ts_forecasting_image} or public health.

After focusing on computer vision and natural language processing, the field of XAI has recently increased its research effort on time series, see \cite{ts_xai_review, rojat_xaisurvey_2021} for a systematic review.
Often, XAI methods originating from other domains like computer vision can be applied to time series classifiers in a straightforward fashion, as they are based on the same architectures like CNNs or RNNs \cite{dl_tsc}. 
Prominently, LRP has been applied to explain time series classifiers in the domain of gait analysis, \cite{gait_2022}, audio classification \cite{audiomnist_2018}, ECG \cite{vij_ecg_2020,west_ecg_2017} and EEG \cite{sturm_eeg_2016}. Other examples of established XAI methods applied to time series are Gradient$\,\times\,$Input (G$\times$I) for ECG data \cite{strodthoff_ecg_2018}, Integrated Gradient for  hydrology \cite{hydrology_lstm_ig}, or \cite{tsc_from_scratch} adopted Grad-CAM for generic time series from the UCR dataset. 
All of these methods produce attribution scores for single time points. However, the explanations cannot rely on the visual interpretability of the single features, i.e. time points, which limits their usability for time series \cite{ts_xai_review}.  This observation is in line with \cite{jeyakumar_how_2020}, who compare XAI methods across input domains (computer vision, natural language processing, and time series) and find that users prefer nearest matching training samples as explanations over input overlayed by relevance scores for time series. In \cite{posthoc_ft_ts_xai}, the authors promote training time-series classifiers in frequency and time-frequency domain in order to make post-hoc explanations by Shapley values or Sobol indices more interpretable. 
By attaching a virtual inspection layer that allows for evaluating relevance scores in an interpretable domain, e.g. the frequency or time-frequency domain, our approach allows renders explanations produced by all aforementioned feature-wise post-hoc XAI methods for classifiers operating in the time domain interpretable.

Recently, a new class of XAI methods, that generate concept-based explanations, emerged \cite{mcd, crp, tcav, chormai22-disentangled, LRPxProtopnet}, that have been applied to hidden feature layers of time series classifiers \cite{crp, timeseries_concepts_II}. While concept-based explanations increase interpretability by contextualizing the explanation with the help of concept prototypes, they still suffer from the limited interpretability of the features they are based on.

Further, there is a multitude of XAI methods specialized in time series: \cite{tsviz} visualizes CNNs by clustering filters and measuring their influence in terms of the gradient, \cite{timeseries_concepts_I} measures the impact of user-defined filters applied in the input space on classification accuracy, \cite{guidotty_anytsc_2020} builds surrogate models based on shapelets, and \cite{ts_counterfactuals} explains the prediction using counterfactual samples from the training set.
These approaches try to improve interpretability by establishing new explanations forms that differ from traditional heatmaps as produced by methods like LRP or IG. The downside of this strategy is that there are no theoretical guarantees for the explanation, like most prominently relevance conservation, which means relevances sum up to the prediction.

Another prominent research area in XAI is the evaluation of explanation techniques. While most evaluation techniques have been developed for general input domains (pixel flipping, localization) \cite{samek_evaluation}, few works address the question of an evaluation specifically for time series. Here, \cite{schlegel_XAITS_2019} proposes a method to evaluate model fidelity of single time point attribution XAI methods. However, we note that so far, there is no existing evaluation techniques available to compare explanations using different units of interpretability (e.g.\ time or frequencies), something which we address in \Cref{sec:feature_flipping} of this paper.

\section{Using LRP to propagate relevance to interpretable representations}

\subsection{Virtual inspection layer}
Let us view a neural network as a composition of functions,
\begin{equation*}
 f(x) = f_L \circ \dots \circ f_1 (x) \,.
\end{equation*}
where each function can be e.g.\ a layer or a block.
We can quantify the relevance $R_f(x_i)$ of each feature $i$ in $x$ towards $y=f(x)$ by a local XAI method. While the representation of the datapoint as $x$ is not interpretable for humans, we assume there is an invertible transformation $\mathcal{T}(x) = \tilde x$, that renders $x$ interpretable. Without the need to retrain the model on $\{ \tilde x\}$, we can now quantify the relevance of $\tilde x_i$ by attaching the inverse transform to the network 
\begin{equation}
 f(x) = f_L \circ \dots \circ f_1 \circ \mathcal{T}^{-1}  \circ \underbrace{\mathcal{T}(x)}_{\tilde x} \, , \label{eq:f1t-1}
\end{equation}
and compute the relevance scores $R_f'(\tilde x_i)$ for the interpretable representation of the data.
In general, an interpretable-representation-inducing bottleneck can be inserted at any layer of the network,
\begin{equation*}
 f( x) = f_L \circ \dots \circ \mathcal{T} \circ \mathcal{T}^{-1} \dots \circ f_1  (x) \, .
\end{equation*}
In the following, we will specialize to DFT regarding $\mathcal{T}$, as our focus in on time series classification, and LRP regarding the local XAI method. 

\subsection{Brief review of LRP}
LRP is a backpropagation-based local XAI method, which decomposes the output of a deep neural network in terms of the input features in a layer-by-layer fashion to arrive at the \textit{relevance} of the input features towards the final prediction. Its central property is the conservation of relevance at each layer.
LRP propagates relevance $R_j$ from layer with neurons $j$ to the layer below with neurons $i$, by summing over all relevances passed from neurons $j$ to neuron $i$,
\begin{equation}
    R_i = \sum_j R_{i \xleftarrow{} j} \, .
\end{equation}
Generically, 
\begin{equation}
    R_{i \xleftarrow{} j} =  \frac{z_{i,j}}{\sum_i z_{i,j}} R_j \,
\end{equation}
where $z_{i,j}$ quantifies how much neuron $i$ contributed towards the activation of neuron $j$, and is usually dependent on the activation $a_i$ and the weight $w_{ij}$ between the neurons.
The sum in the denominator ensures the conservation property $\sum_i R_i = \sum_j R_j$.
There are numerous choices for $z_{i,j}$ corresponding to propagation rules. Which rules to choose depends on the model under consideration (see e.g.\ \cite{KohIJCNN20, lrp_overview}
To summarize, LRP propagates relevance scores $R _j$ at layer $j$ onto neurons of the lower layer $i$ by the rule,
\begin{equation} \label{eq:lrp}
	R_i = \sum_j  \frac{z_{i,j}}{\sum_i z_{i,j}} R_j  \,
\end{equation}
until the input layer is reached.

\subsection{Relevance Propagation for the Discrete Fourier Transformation}
For a neural network trained in time domain, we can employ \Cref{eq:lrp} to quantify the relevance of each time step towards the prediction. Here, we lay out how to propagate relevance one step further into the frequency domain.
A signal in time domain $x_n$, $n=0,...,N-1$ is connected to its representation in frequency domain $y_k \in \mathcal{C}$, $k=0,...,N-1$, via the DFT. The DFT and its inverse are simply linear transformations with complex weights,
\begin{align}
	y_k 
		&= \dft(\{x_n\}) = \frac{1}{\sqrt{N}} \sum_{n=0}^{N-1} x_n \left [ \cos (\frac{2\pi k}{N}) - i \sin (\frac{2\pi k}{N}) \right ] \label{eq:DFT}\\  
	x_n	&= 	\idft(\{y_k\})=  \frac{1}{\sqrt{N}} \sum_{k=0}^{N-1} y_k \left [ \cos (\frac{2\pi k}{N}) + i \sin (\frac{2\pi k}{N}) \right ]  \,. \label{eq:iDFT}
\end{align}
We require relevances of $y_k$ to be real. Thus, we proceed by writing the signal in frequency domain as a concatenation of real and imaginary parts, \linebreak \mbox{$[\real{y_0}, \real{y_1},\dots,\real{y_{N-1}},\dots, \imag{y_1},\dots,\imag{y_{N-1}}]$}.
As visualized in \Cref{fig:scheme}, we attach a layer that performs the inverse DFT in \cref{eq:iDFT} to the model, before the first layer $f_1$ that operates on the signal in time domain. 
For real signals $x_n \in \mathcal{R}$ we can express the inverse DFT as,
\begin{equation}
	x_n = \frac{1}{\sqrt{N}} \sum_{k=0}^{N-1} \real(y_k)  \cos\Bigl(\frac{2\pi }{N nk}\Bigr) - \imag(y_k)  \sin\Bigl(\frac{2\pi }{N nk}\Bigr) \, . \label{eq:iDFTlayer}
\end{equation}
We assume that relevance values $R_n$ for $x_n$ are available and that they are of form $R_n = x_n c_n$ (a property ensured by most LRP rules, in particular, LRP-$0/\epsilon/\gamma$). 
Now, the question is how to transform relevance in time domain $R_n$ to relevance in frequency domain. For LRP, the first question is how much each frequency component $\real(y_k), \imag(y_k)$ contributes to each time point $x_n$, i.e. finding an expression for $z_{i,j}$ in \Cref{eq:lrp}.
The inverse DFT in \Cref{eq:iDFTlayer} is only a homogeneous linear model, i.e. of type $f(x) = wx$. 
Thus, the contribution of neuron $\real(y_k)$, $\imag(y_k)$ to $x_n$ is the value of the neuron itself times the weight,
\begin{align}
    z_{k,\real,n} &= \real(y_k) \cos(\frac{2\pi nk}{N}),\\
    z_{k,\imag,n} &= - \imag(y_k) \sin(\frac{2\pi nk}{N})\, .
\end{align}
In fact, it can easily be shown, that this is the value popular local XAI methods LRP-0, Deep Taylor Decomposition, Integrated Gradients, PredDiff, and Shapley values default to for homogeneous linear transformations if one sets the respective reference value to zero \cite{samek_XAI_2021}.

Now, we apply \cref{eq:lrp} to aggregate the contributions of each neuron $R_{k,\real}$, $R_{k,\imag}$ towards the model output and find,
\begin{align} 
    \label{eq:rel_real_imag}
	R_{k,\real} &= \real(y_k) \sum_n \cos(\frac{2 \pi k n}{N}) \frac{R_n }{x_n }  \\
	R_{k,\imag} &= - \imag(y_k) \sum_n \sin(\frac{2 \pi k n}{N}) \frac{R_n}{x_n } \,.
\end{align}
Here, we assume $R_k=0$ if $x_k=0$ and define $0/0=0$. In practice, we add a small-valued constant $\epsilon$ to the denominator for numerical stability.

Now, leveraging addivity of LRP attributions, we define $R_{k} = R_{k,\real} + R_{k,\imag}$. To abbreviate the form of the sum, we separate $y_k$ into amplitude $r_k$ and phase $\varphi_k$, i.e. $\real(y_k) = r_k \cdot \cos(\varphi_k)$ and $\imag(y_k) = r_k \cdot \sin(\varphi_k)$, and find,
\begin{align}
\boxed{
	R_k
	=  r_k \sum_n \cos(\frac{2 \pi k n}{N} - \varphi_{k})  \frac{R_n}{x_n}} \, .  \label{eq:dftlrp}
\end{align}

\subsection{Relevance Propagation for the Short-time Discrete Fourier Transformation} \label{sec:stdft}
For slowly varying, quasi-stationary time series like audio signals, one is interested in how the frequency content varies over time. Here, one applies the short-time DFT (STDFT) which, connects the signal in time to the time-frequency domain.
For the STDFT, one computes the DFT of potentially overlapping windowed parts
of the signal \cite{stdft},
\begin{equation} \label{eq:stdft}
   v_{m,k} = \dft(\underbrace{x_n \cdot w_m(n)}_{s{m,n}})\, ,
\end{equation}
where $w_m(n)$ is a window function with window width $H$, that selects the segment of the signal to be analyzed. It is shifted by $m\cdot D$ time points. To sequentially cover the whole signal, we require $0<D\leq H$ for the shift length.
To recover the original signal $\{x_n\}$ given $\{S_{m,n}\}$, first, we compute the inverse DFT in \cref{eq:iDFT} of $\{v_{m,k}\}$ to obtain $\{s_{m,n}\}$. Second, we rescale $\{s_{m,n}\}$ by the sum over the windows $w_m(n)$ over shifts $m$ to obtain $\tilde x_n$:
\begin{equation}
    \tilde x_n = \frac{\sum_m \idft(\{v_{m,k}\}) } {\sum_m w_m(n)} \, . \label{eq:iSTDFT}
\end{equation}
This so-called weighted overlap-add (WOLA) technique imposes only a mild condition on the windows $w_m(n)$ for perfect reconstruction $\tilde x_n=x_n$, which is,
\begin{equation*}
    \sum_m w_m(n) \neq 0  \;\; \forall n  \,.
\end{equation*}
In the following, we write $W_n = \sum_m w_m(n)$. In \Cref{sec:COLA}, we show an alternative formulation of the inverse STDFT, which imposes stricter conditions on the windows.
Analogous to \Cref{{eq:rel_real_imag}}, we propagate the relevance $R(x_n)$ to the real $\real(z_{mk})= r_{m,k} \cos(\varphi_{m,k})$, and imaginary part $\imag(z_{mk})= r_{m,k} \sin(\varphi_{m,k})$ of $z_{mk}$,  
\begin{align*}
    R_{m,k,\real} &= r_{m,k} \cos(\varphi_{m,k}) \sum_n \cos(\frac{2 \pi k n}{N}) \cdot  W_n^{-1} \frac{R_n}{x_n } \\
    R_{m,k,\imag} &= r_{m,k} \sin(\varphi_{m,k}) \sum_n \sin(\frac{2 \pi k n}{N}) \cdot W_n^{-1} \frac{R_n}{x_n} \, .
\end{align*}
Aggregating the relevance of real and imaginary part yields,
\begin{equation}
    \label{eq:stdft-lrp}
    \boxed{
    R_{m,k} = r_{m,k} \sum_n \cos(\frac{2 \pi k n}{N} - \varphi_{m,k}) \cdot  W_n^{-1} \frac{R_n}{x_n}
    }\, .
\end{equation}

We now specialize to an appropriate choice for the window function $w_m(n)$. In general, any spectrum obtained from a windowed signal suffers from spectral leakage. The DFT of the product between the signal in time domain and the window function is the convolution between the DFT of the original signal and the DFT of the windowing function. Thus, the latter introduces new frequency components, known as spectral leakage.\footnote{In fact, this is inevitable for the DFT of any signal, not just for STDFT, because a discrete and finite signal is always subject to sampling and windowing.} Depending on the shape of the windowing function, spectral leakage can cause two opposing issues. On the one hand, it can restrict the ability to resolve frequencies that are very close but have a similar amplitude (\textit{low resolution}). On the other hand, it can limit the ability to resolve frequencies that are far apart from each other but have dissimilar frequencies (\textit{low dynamic range}). Windows with a rectangular shape have a high resolution but a low dynamic range. On the other end of the spectrum, windows with much more moderate changes on the edges like the half-sine window have a high dynamic range but a low resolution.
At this point, the window function and shift can be chosen according to the requirements of the time series at hand. 

\subsection{Properties of DFT-LRP} \label{sec:properties}
Here, we list the conservation and symmetry properties of (ST)DFT-LRP, inherited from LRP and DFT.

\textbf{(1) Total relevance conservation} The total relevance in frequency domain equals the total relevance in time domain, i.e. $\sum_k R_k = \sum_n R_n$. This is easily validated by checking
\begin{equation}
    \sum_k R_k =  \sum_n \underbrace{\sum_k r_k \cos(\frac{2 \pi k n}{N} - \varphi_{k})}_{x_n}  \frac{R_n}{x_n} \, , \nonumber
\end{equation}
for DFT-LRP and
\begin{align*}
     \sum_{k,m} R_{k,m} &=  \sum_m  \sum_n \underbrace{\sum_k r_{m,k}  \cos(\frac{2 \pi k n}{N} - \varphi_{m,k}) \cdot  W_n^{-1}}_{x_n \cdot I_{n \in m}} \frac{R_n}{x_n} \\ &=  \sum_m \sum_{n \in m} R_n = \sum_{n} R_n \,,
\end{align*}
for STDFT-LRP. In particular, due to the rescaling with $W^{-1}$, this is given for any window choice and overlap.

\textbf{(2) Relevance conservation in time bins} In time-frequency domain, we might require more fine-grained relevance conservation over time bins in some settings. Here, we want to obtain the total relevance over time interval $n \in m$ in time domain when we sum over frequency bins in time bin $m$ in the time-frequency domain, i.e. $\sum_{k} R_{k,m} = \sum_{n \in m} R_n$.

In the case of overlapping windows with shift $D<H$, the signal is stretched in time domain and there is no clear assignment between relevance in time bins in time and time-frequency domain. Thus, we can assign this property to STDFT-LRP only when $D=H$.
This singles out the rectangular window, because windows with smoothed edges and no overlap suffer from information loss at the edges where the signal receives weights close to zero when $D=H$. 
To summarize, when we require fine-grained relevance conservation over time bins, like in \Cref{sec:sythetic_task}, we need to restrict to the rectangular window with shift $D=H$.

\textbf{(3) Symmetry} We  assume $x_n \in \mathbb{R}$, which implies that the spectrum is even symmetric $y_k = y_{-k\,mod\,N}$. As can be read from \Cref{eq:dftlrp}, this is also true for $R_k$. This symmetry can be leveraged for reduced computational cost, as one only needs to evaluate $R_k$ only for $k \in [0,N/2+1]$.

\section{Results}
First, we empirically evaluate our method on a synthetic dataset with ground-truth annotations in \Cref{sec:sythetic_task} and on a real-world dataset via feature flipping in \Cref{sec:sythetic_task}. Next, we demonstrate the utility of our approach in two use-cases: We compare the strategies of two audio classifiers trained on different input domains in \Cref{sec:use_case_1} and show how DFT-LRP reveals Clever hans strategies of an audio and ECG-classifier in \Cref{sec:use_case_2}

\subsection{Datasets and models} \label{sec:datasets}
 We list all datasets and models that we use in the following sections.
\emph{Synthetic} The signal is a simple superposition $M$ sinusoids,
\begin{equation*}
    x_n = \sum_j^M a_j \cdot \sin (\frac{2\pi n}{N k_j} + \varphi_j) + \sigma y
\end{equation*}
with amplitude $a_j$, frequency $2\pi/N k_j$, random phase $\varphi_j$, and additive Gaussian noise $y \sim \mathcal{N}(0,1)$ with strength $\sigma$.  We choose the signal length as $N=2560$ and restrict to $0<k_j<60$. The task is to detect a combination of one to four frequencies from the set $k^* = \{k_1,k_2,k_3,k_4\}$ in the time representation of the signal. Here, each combination of $\{k_i\}$ from the powerset of $k^*$, i.e. $\{\}, \{k_1\},  \dots, \{k_1,k_2\}, \dots, \{k_1,k_2,k_3\}, \dots, \{k_1,k_2,k_3,k_4\}$, corresponds to a label. We choose $k^* = \{5,16,32,53\}$ for the set. We train a simple Multi-Layer-Perceptron model with two hidden layers and ReLU activation on $10^6$ samples on a \textit{baseline} task with noise strength $\sigma=0.01$ and a \textit{noisy} task with $\sigma=0.8$. The model reaches a test set accuracy of 99.9\% and 99.7\%, respectively.

\emph{AudioMNIST} This dataset by \cite{audiomnist_2018} consists of 3000 recordings of spoken digits (0-9) in English with 50 repetitions of each digit by each of 60 speakers. Besides the actual spoken digit, the dataset contains meta-information such as biological sex and accent of all speakers.
Following \cite{audiomnist_2018}, we down-sample recordings from 16kHz to 8kHz and zero-pad them, such that each recording is represented by a vector of length 8000.
We train the same 1d CNN classifier as in \cite{audiomnist_2018} on the raw waveforms and achieve an accuracy of 92\% on 96\% on the sex and digit classification task, respectively.

\emph{MIT-BIH}
The ECG arrhythmia database by \cite{mit-bih} consists of ECG recordings from 47 subjects, with a sampling rate of 360\;Hz. The preprocessing follows \cite{ecg_transferable_representation}, who isolated the ECG lead II data, split and padded the data into single beats with a fixed length of 1500\;ms at a sampling rate of 125\;Hz. At least two cardiologists have annotated each beat and grouped the annotations into five beat categories: 1) normal beats etc., 2) supraventricular premature beats, etc, 3) premature ventricular contraction and ventricular escape, 4) fusion of ventricular and normal, and 5) paced/ unclassifiable, etc. in accordance with the AAMI EC57 standard.  
The model under consideration is a 1-dim. CNN with three convolutional layers and a classification head consisting of three dense layers, all with ReLU activations, that classifies an ECG signal in time domain into five beat categories with an  
accuracy of 95.3\%.

\subsection{Evaluation on synthetic data with ground truth} \label{sec:sythetic_task}
We evaluate (ST)DFT-LRP in a setting where ground truth relevance attributions in frequency and time-frequency domain are available for a simple task on synthetic data. First, we quantitatively evaluate how well (ST)DFT-LRP explanations and explanations of attribution methods equipped with a virtual DFT layer align with the ground truth. Second, we qualitatively evaluate the interpretability of explanations in time versus frequency domain.

\subsubsection{Quantitative evaluation} \label{sec:sythetic_task:quant}
We base this evaluation on explanations of the frequency detection models trained on the \textit{baseline} and \textit{noisy} task for the respective test split of the synthetic dataset described in \Cref{sec:datasets}.
We compute LRP relevances using the $\epsilon$-rule in time domain and apply (ST)DFT-LRP according to \Cref{eq:dftlrp} and \Cref{eq:stdft-lrp} to transform them to frequency and time-frequency domain. We compare to other local attribution methods, namely Sensitivity \cite{morch1995visualization, baehrens2010explain}, gradient times input (G$\times$I) (e.g.\ \cite{DBLP:conf/iclr/AnconaCO018}), and Integrated Gradient (IG) \cite{sund_ig_2017} that we equip with a virtual inspection layer. To this end, we attach an inverse (ST)DFT layer according to \Cref{eq:iDFTlayer} and \Cref{eq:iSTDFT} to the input layer like in \Cref{fig:scheme}. Then, we perform the attribution method for the new model that takes the signal in frequency (time-frequency) domain (split into real and imaginary part) as input. For IG, we use $x_n=y_k=z_{mk}=0$ as a baseline. For all attribution methods, we employ the zennit package \cite{anders2021software}.
Given the simplicity of the task and the high test set accuracy of close to 100\%, we can assume that ground-truth explanations correspond to attributing positive relevance only to the subset of $k^*$ related to the respective label.
To quantitatively evaluate how well the explanations align with this ground truth, we define a \textit{relevance localization score} $\lambda$, 
\begin{equation} \label{eq:rel_loc}
    \lambda = \sum_{k \in k^*} R_k / \sum_k R_k I_{R_k>0}  \, ,
\end{equation}
which measures the ratio of the positive relevance that is attributed to the informative features $\{k_i\}$. A high $\lambda$ corresponds to accurate relevances in frequency or time-frequency domain. 

\begin{table*}[]
\centering
\small
\caption{Positive relevance localization $\lambda$ of explanations in the frequency and time-frequency domain for synthetic frequency detection tasks with low (baseline) and high (noisy) additive noise. Relevances from LRP, IG, and G$\times$I all show equal localization scores. The error is below $0.01$ in all cases.}
\begin{tabular}{lcccc}
\toprule
task & \multicolumn{2}{l}{baseline} & \multicolumn{2}{l}{noisy} \\
method &      \{LRP, IG, G$\times$I\} & Sens. &   \{LRP, IG, G$\times$I\}  & Sens. \\
\midrule
$\lambda_{DFT}$     &     0.94 &        0.51 &  0.80 &        0.46 \\
$\lambda_{STDFT-N/10}$ &     0.36 &        0.51 &  0.29 &        0.46 \\
$\lambda_{STDFT-N/4}$ &     0.67 &        0.51 &  0.55 &        0.46 \\
$\lambda_{STDFT-N/2}$   &     0.80 &        0.51 &  0.67 &        0.46 \\
\bottomrule
\end{tabular}
\label{tab:dft_rel_loc}
\end{table*}

In \Cref{tab:dft_rel_loc}, we show the mean $\lambda_{\text{DFT}}$ and $\lambda_{\text{STDFT}}$ for heatmaps in frequency and time-frequency domain across $1000$ test set samples. In time-frequency domain, we evaluate $\lambda$ for STDFTs with window widths $D=N/10,N/4,N/2$.

Because the simple MLP model with only two hidden layers and ReLU activation is only slightly non-linear, and LRP, G$\times$I, and IG reduce to the same attribution for a linear model \cite{samek_XAI_2021}, the attributions among these methods are very similar, resulting in equal relevance localization scores $\lambda$.

For the \textit{baseline} task, we observe almost perfect relevance localization $\lambda_{DFT}$ in the frequency domain for DFT-LRP and equivalent methods. For the \textit{noisy} task, $\lambda_{DFT}$ reduces to 0.80. When cutting the sum in \Cref{eq:rel_loc} at the maximum frequency of the signal ($k=60$), this gap disappears, revealing that DFT-LRP and equivalent methods mix a small part of the total relevance with noise. 

Further, we observe that $\lambda_{STDFT}$ is generally lower than $\lambda_{DFT}$ and increases with higher window width $D$ for STDFT-LRP and equivalent. This is due to the time-frequency resolution trade-off inherent to STDFT. The higher $D$, the higher the frequency resolution, i.e. the ability to resolve similar frequencies, and the lower the time resolution. Because the signal is stationary, $\lambda_{STDFT}$ is affected only by the increase in frequency resolution, not the decrease in time resolution. 

Lastly, sensitivity mostly shows much lower localization scores $\lambda$ than LRP and other methods. This is because it looks at local effects instead of overall feature contributions. Only for $\lambda_{STDFT-N/10}$, sensitivity has the highest score. This is because sensitivity takes into account only the gradient, not the signal itself determines the sensitivity, and the weights in the inverse STDFT layer are the same for each window shift as for the DFT layer. Thus, $\lambda_{DFT}$ and $\lambda_{STDFT}$ are equal and STDFT-sensitivity does not suffer from the limited frequency resolution.

In summary, DFT-LRP is superior to sensitivity and equivalent to IG and G$\times$I for this simple task. DFT-LRP and STDFT-LRP can reliably recover ground-truth explanations, up to the slight mixing of relevance with noise and limitations inherent to STDFT, i.e. the time-frequency resolution trade-off.

\subsubsection{Qualitative evaluation}
\begin{figure}[th]
	\center
    \hspace{1.5cm}\small{time}  \hspace{3cm}\small{frequency} \\
    \rotatebox{90}{\hspace{1cm}\small{signal}} 
	\includegraphics[width=0.4\linewidth]{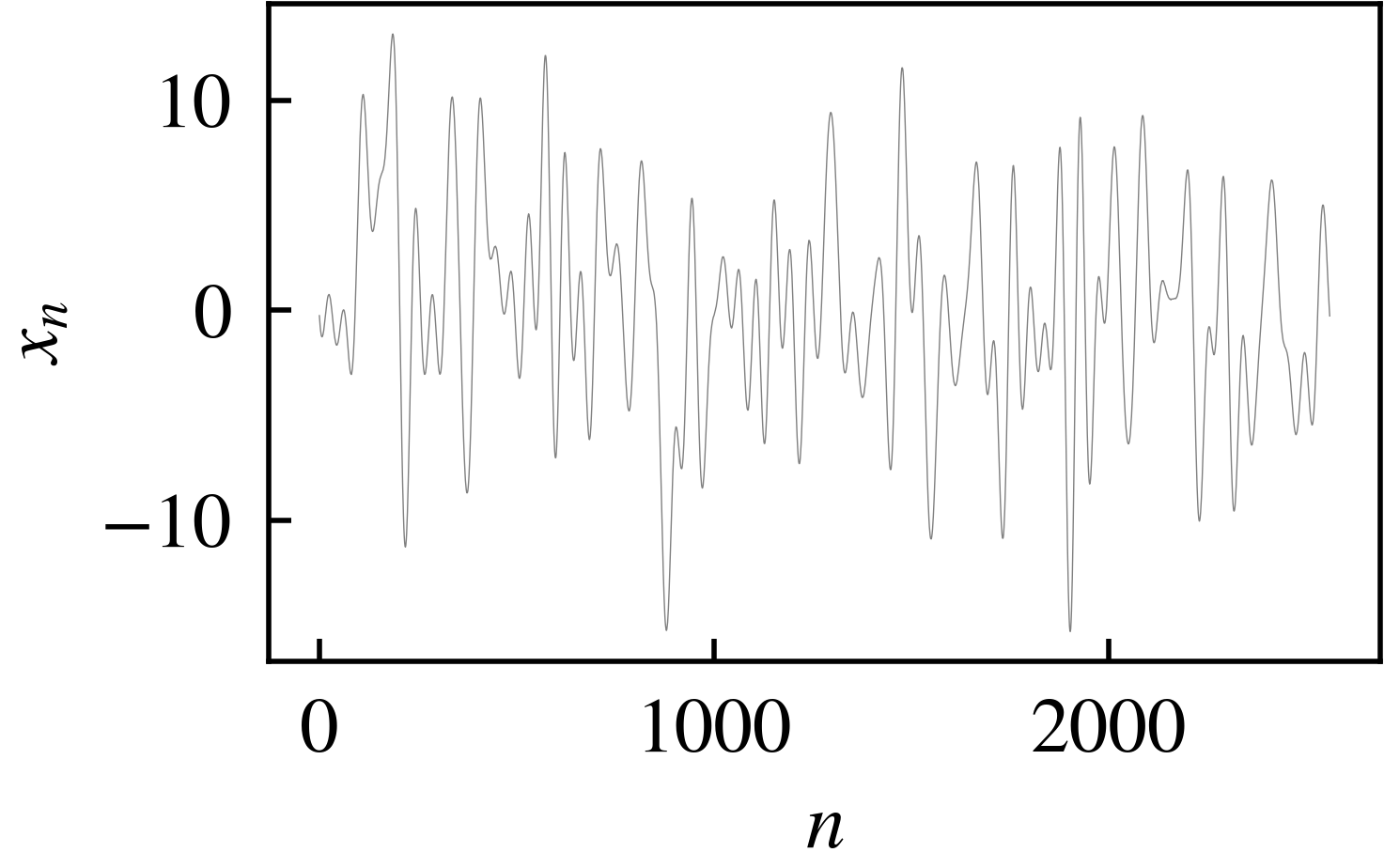} \hspace{0.2cm}
    \includegraphics[width=0.4\linewidth]{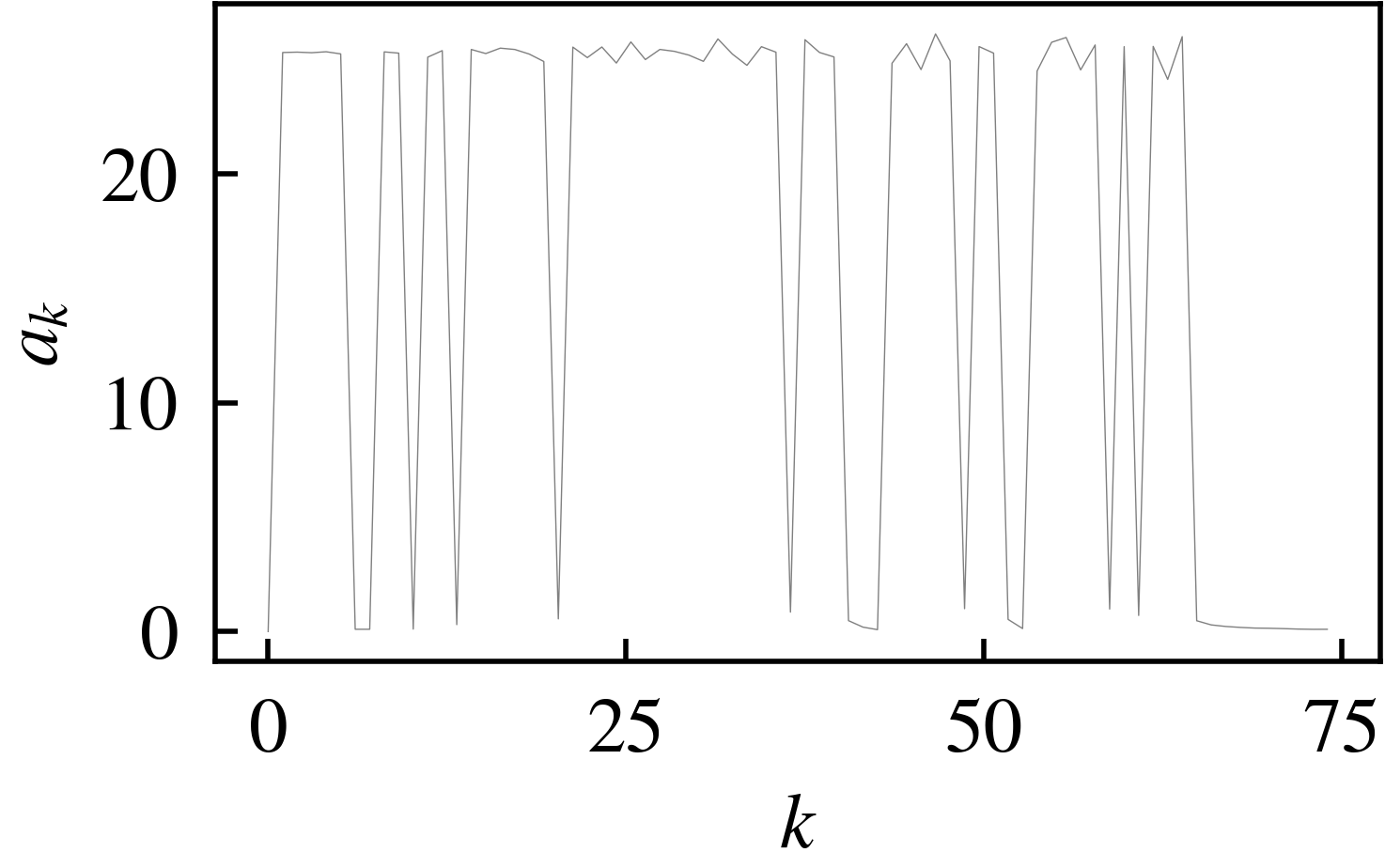}\\
    \vspace{0.2cm}
    \rotatebox{90}{\hspace{1cm}\small{relevance}} 
    \includegraphics[width=0.4\linewidth]{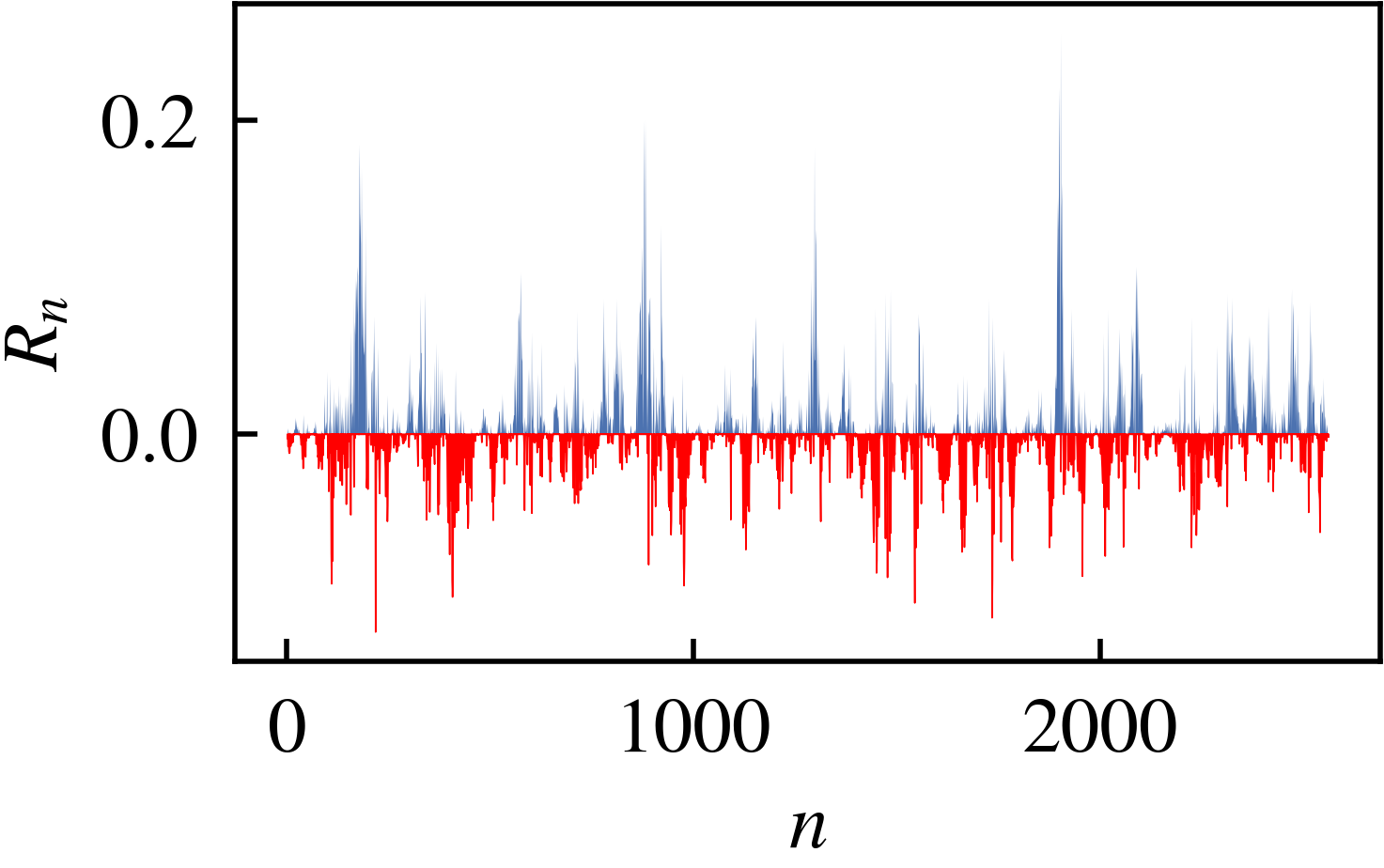} \hspace{0.2cm}
    \includegraphics[width=0.4\linewidth]{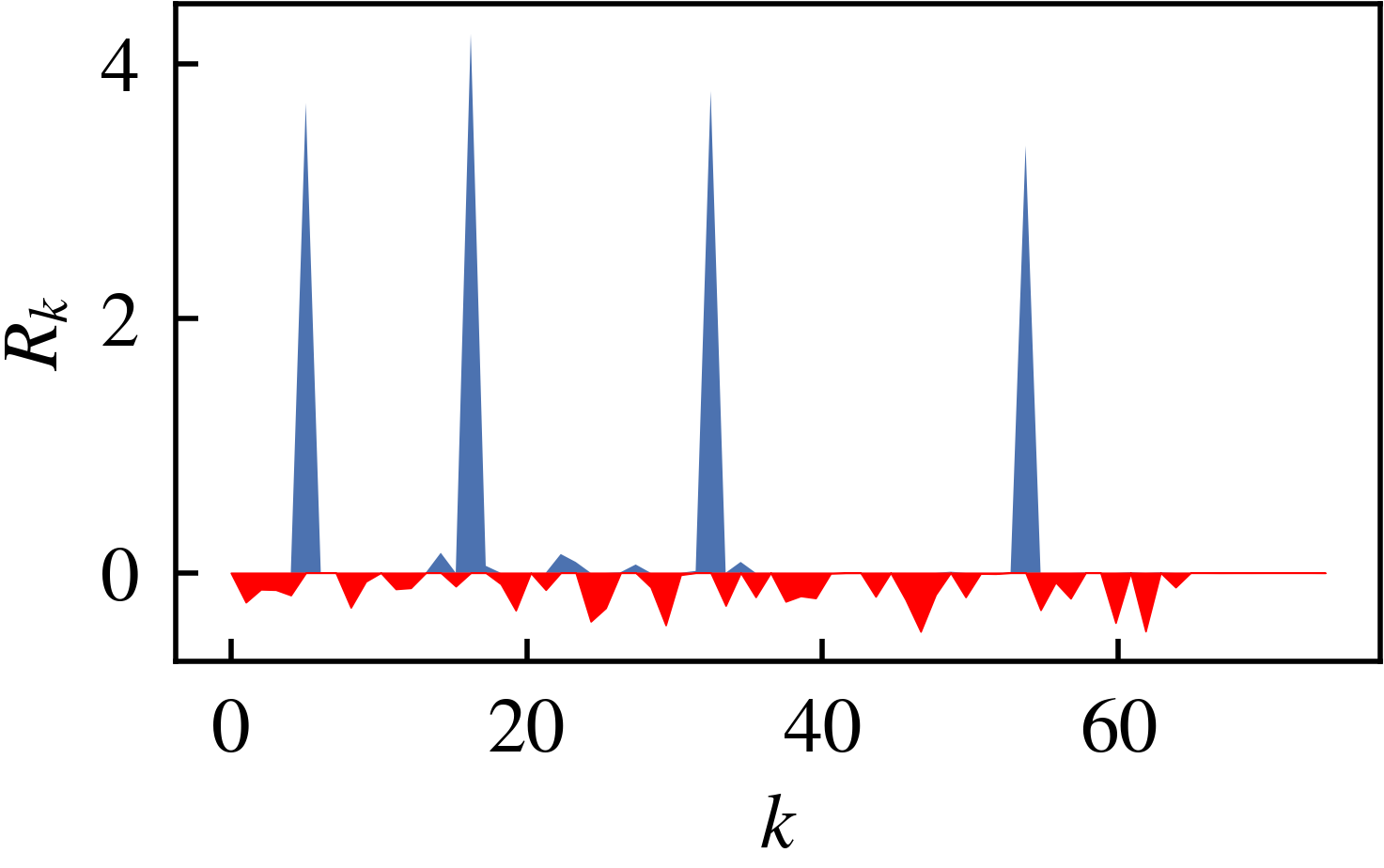}    
	\caption{Time and frequency signal (first row) and relevances (second row) for the frequency detection task on the synthetic data. Relevances are based on the LRP-$\epsilon$ rule and (ST)DFT-LRP. Relevance in time domain is distributed rather uniformly in time domain, but is clearly localized on the frequencies to detect, i.e. $\{k_1,k_2,k_3,k,4\}$, in frequency domain, revealing the classifier strategy.
 }
	\label{fig:sytnhetic_proofofconcept}
\end{figure}

We briefly demonstrate the advantage of relevance propagation to frequency domain.
We show (ST)DFT-LRP relevances in time and frequency domain in \Cref{fig:sytnhetic_proofofconcept} for the baseline task and a signal corresponding to the label $\{5,16,32,53\}$. 
In the domain, relevance is distributed rather uniformly over the entire signal. In contrast, relevance is clearly localized on the ground-truth informative frequencies $k^*$ frequency domain. 
We argue, that the classifier strategy is only comprehensible after relevance propagation to frequency domain.

\subsection{Evaluation on real-world data} \label{sec:feature_flipping}
For real-world audio data, we 1) test which feature domain -- time, frequency or time-frequency -- is the most \textit{informative} to the model across different XAI methods, and 2) compare the \textit{faithfulness} of different XAI methods in each feature domain. 
To this end, we measure the complexity of heatmaps \cite{HedJMLR23} and perform feature flipping experiments, analogous to Pixel-flipping, which is often deployed to benchmark XAI methods in computer vision \cite{samek_evaluation}.

We base our evaluation on the digit classification model trained on the AudioMNIST dataset.
Again, we consider LRP, IG, G$\times$I, and sensitivity. We compute LRP relevances in time domain by applying the $z^+$-rule to convolutional and the $\epsilon$-rule to dense layers, and then apply ST(DFT)-LRP via \Cref{eq:iDFTlayer} and \Cref{eq:iSTDFT} to propagate relevances $R_n$ from time domain $x_n$ to frequency $y_{k}$ and time-frequency $v_{m,k}$ domain.
Relevance scores for sensitivity, G$\times$I, and IG in all domains are computed like in the previous section, i.e. by attaching an inverse Fourier layer to the original input layer as a virtual inspection layer. We choose a rectangular window of size $H=N/10$ and hop length $D=H$ for the STDFT.\footnote{We choose a rectangular window, so we can choose the hop size to equal the window width, in order to not introduce artifacts by flipping a time-frequency feature that overlaps with another feature in time.}

First, we evaluate the complexity of the heatmaps by measuring their Shannon entropy. In the domain which is the most \textit{informative} to the model, a few features are enough to make the prediction, thus relevance concentrates on them and heatmaps have low complexity.
Second, we perform feature flipping in time, frequency, and time-frequency domain by either flipping the respective features to a zero baseline in order of their relevance scores (smallest destroying feature, SDF) or starting with an empty signal and adding the most relevant features first (smallest constructing feature, SCF). After each feature modification, i.e. addition or deletion, we measure the model's output probability for the true class. For frequency and time-frequency, we set the amplitude of $y_k, z_{k,m}$ to zero for $k=0,\dots N/2$, considering that the signal is symmetric in frequency domain. In time domain, we set the time point $x_n$ to zero. For comparability of the feature flipping curve across domains, we scale it to the ratio of modified features, where 100\% correspond to $N$ features flipped/added in time, $N/2$ features in frequency and $N/H\cdot N/2$ features in time-frequency domain.
To reduce the results to a scalar score, we compute the area under the curve (AUC) of the feature flipping curves.
A relevance attribution method that is \textit{faithful} to the model reflects in a steep descent/ascent in true class probabilities after flipping/adding the truthfully as most important annotated features.

\begin{figure}[t]
	\center
	\includegraphics[width=\linewidth]{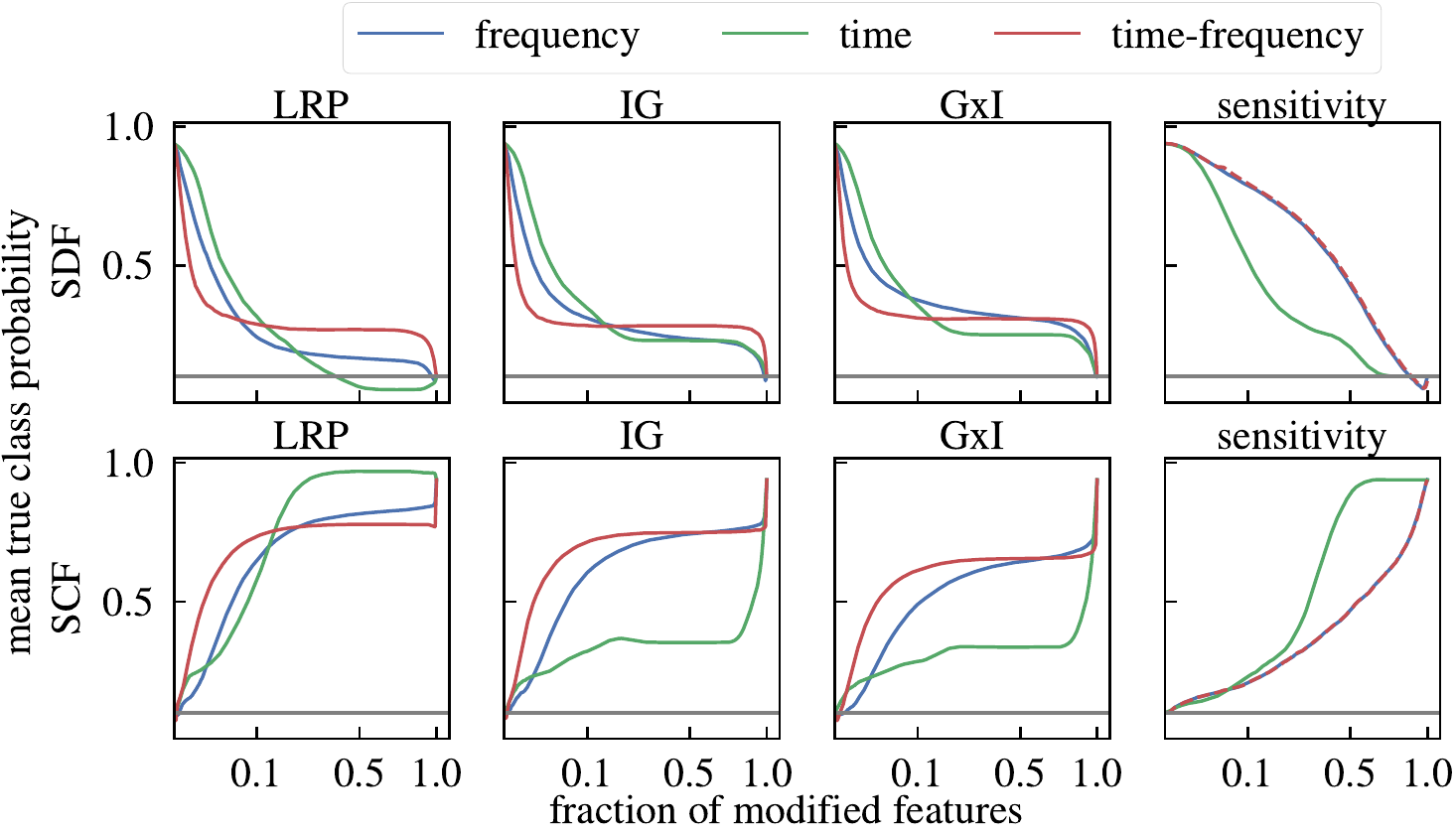}
	\caption{Evaluation of LRP, IG, G$\times$I and sensitivity relevances for a model trained on the AudioMNIST digit classification task via feature flipping: Mean true class probability after feature deletion (SDF) and feature addition (SCF) in time ($x_n$), frequency ($y_k$) and time-frequency ($v_{m,k}$) domain. The horizontal axis is square root scaled. The grey horizontal line corresponds to the chance level, i.e. a probability of 0.1. 
    }
	\label{fig:feature_flipping}
\end{figure}

\begin{figure*}[t]
	\center
    \hspace{1.8cm}\small{time}  \hspace{3.2cm}\small{frequency} \hspace{2.5cm}\small{time-frequency} \\
    \rotatebox{90}{\hspace{1cm}\small{signal}} 
	\includegraphics[width=0.2\linewidth]{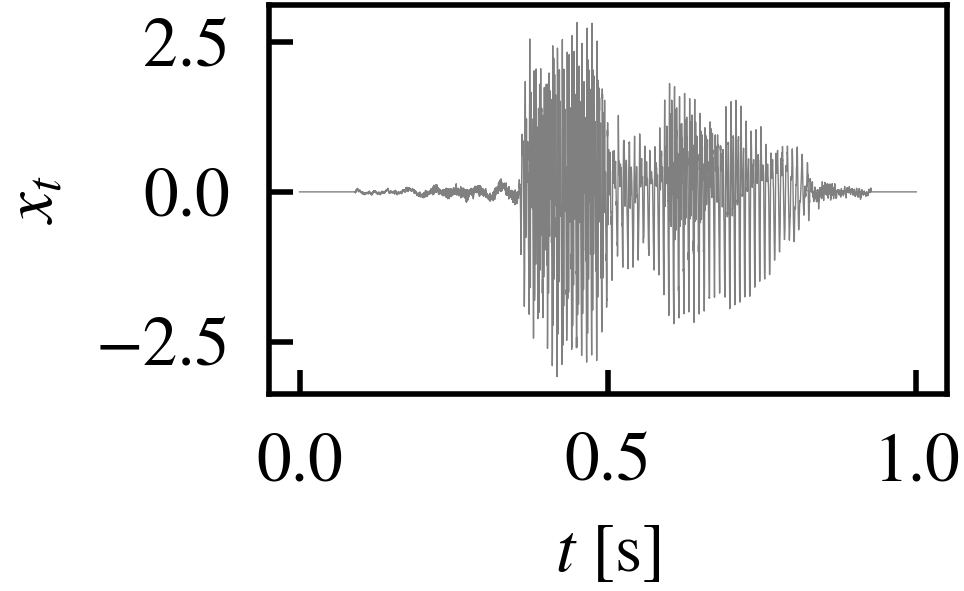} \hspace{0.2cm}
    \includegraphics[width=0.2\linewidth]{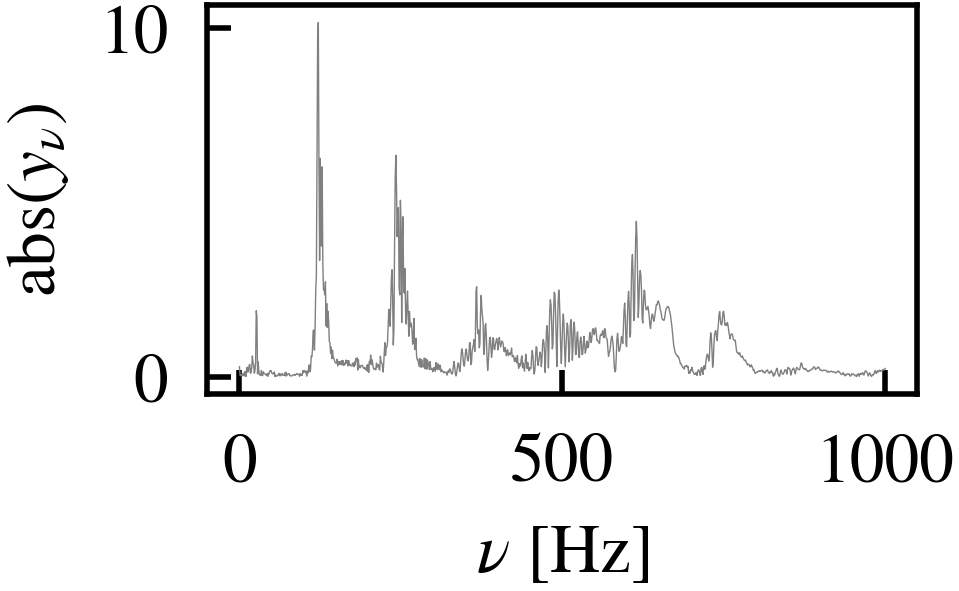}\hspace{0.2cm}
    \includegraphics[width=0.2\linewidth]{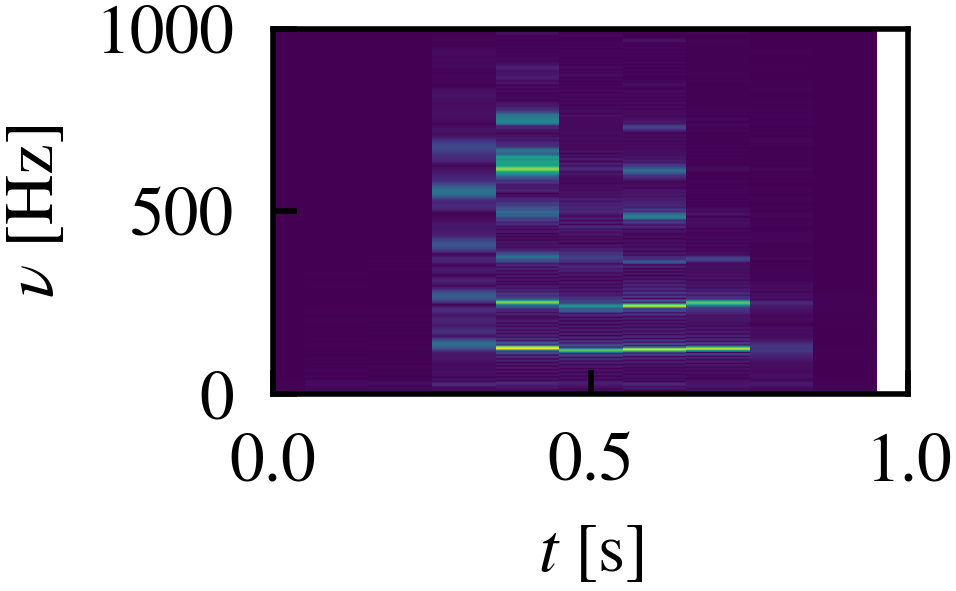}\\
    \vspace{0.2cm}
    \rotatebox{90}{\hspace{1cm}\small{relevance}} 
	\includegraphics[width=0.2\linewidth]{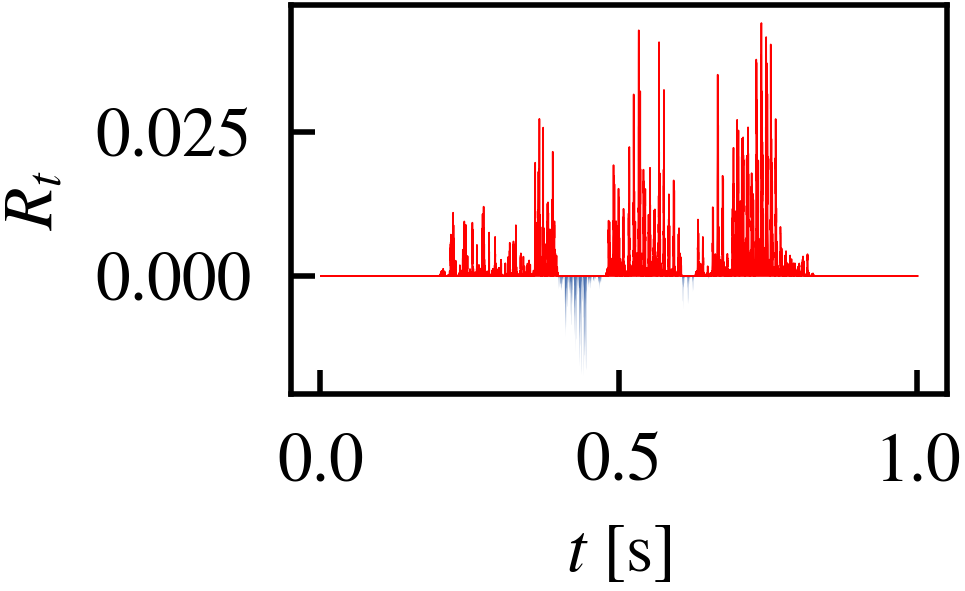} \hspace{0.2cm}
    \includegraphics[width=0.2\linewidth]{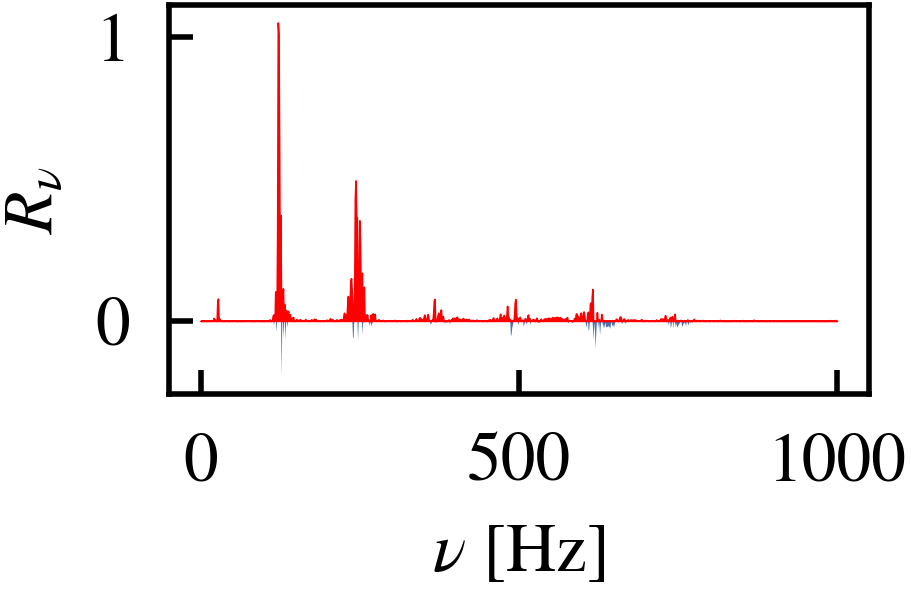}\hspace{0.2cm}
    \includegraphics[width=0.2\linewidth]{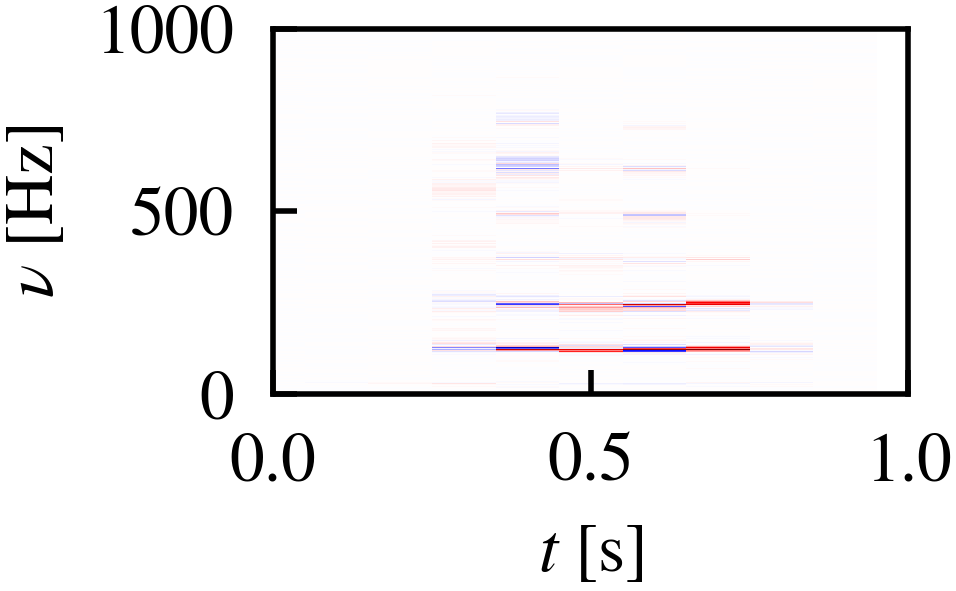}\\  
	\caption{Time, time-frequency, and frequency signal (first row) and relevances  (second row) for the digit detection task on the AudioMNIST data. The signal corresponds to a spoken seven. Relevances are based on the LRP-$z^+$-rule for convolutional and LRP-$\epsilon$ rule for dense layers and (ST)DFT-LRP. Relevance in time domain is distributed rather uniformly over the signal in time domain but is more localized in frequency and time-frequency domain.}
	\label{fig:audiomnist_digit_example}
\end{figure*}

In \Cref{fig:feature_flipping} we show the true class probability against the ratio of deleted/added features, i.e. for SDF and SCF respectively, 
for all attribution methods and input domains. 
We list the corresponding AUC scores of the feature flipping curves and the mean complexity over all heatmaps in \Cref{tab:audiomnist}. 
For a qualitative comparison of explanations across feature domains, we show LRP heatmaps for each domain in \Cref{fig:audiomnist_digit_example} for a randomly selected sample correctly classified as a seven. 

Now, we turn to the question of which feature domain is the most \textit{informative} to the model and compare complexity scores across domains for each XAI method. 
For each method except Sensitivity, the frequency domain shows the lowest complexity, i.e. is the most informative the to model, followed by time and time-frequency domain. However, the visual impression of the heatmaps in \Cref{fig:feature_flipping} contradicts this ranking, as relevance shows distinct peaks at certain frequencies in frequency \textit{and} time-frequency domain, but is distributed rather uniformly in time domain. Thus, we suspect that the higher complexity of time-frequency heatmaps compared to time domain might result from the fringes in the spectrum, produced by the sharp edges of the rectangular window. 
We note, that we cannot accuratly compare informativeness via feature flipping, because here, probability decrease/increase might not only result from deducting/adding information, but also from off-manifold evaluation of the model as setting features to the baseline causes unknown artifacts in the sample and this effect might differ between feature domains. Still, if we focus on the first part of the SDF and SCF feature flipping curves in \Cref{fig:feature_flipping}, where the least artifacts exist, the initial steep decrease/increase in time-frequency and frequency domain supports our findings. 

Again, sensitivity makes no difference between frequency and time-frequency features because it only takes into account the gradient, i.e. the weights of the Fourier transformation, as already described in \Cref{sec:sythetic_task:quant}.
We would like to stress, that the informativeness ranking of input domains is dependent on the quality of the time series to classify, e.g. for time series with time-localized characteristics, time domain might be more informative to the model than frequency domain. Our method novelly enables the comparison between the domains and allows analyzing the model strategy in the most interpretable domain.

Lastly, we compare the faithfulness between XAI methods in each domain in terms of the feature flipping results.
For all domains, ((ST)DFT)-LRP delivers the most \textit{faithful} relevance heatmaps, followed by IG, G$\times$I, and Sensitivity, according to both, SCF and SCF AUC scores. 

\begin{table*}[h]
    \centering
    \small
        \caption{Evaluation of LRP, IG, G$\times$I and sensitivity relevances for a model trained on the AudioMNIST digit classification task: AUC of feature flipping curves for adding (SCF) and deleting (SDF) features in order of their relevance, and complexity scores. The method with the globally highest faithfulness, i.e. highest ($\uparrow$) AUC for SCF and lowest ($\downarrow$) AUC for SDF, is marked in bold. Further, the domain with the lowest complexity is marked in bold for each attribution method. 
        The AUC scores correspond to the feature flipping curves in \Cref{fig:feature_flipping}, where the horizontal axis is square root scaled.
        }
\begin{tabular}{llrrr}
\toprule
            &    &   SCF ($\uparrow$) &   SDF ($\downarrow$) &  complexity ($\downarrow$) \\
method & domain &       &       &             \\
\midrule
\textbf{LRP} & frequency &  0.66 &  0.28 &        \textbf{6.00} \\
            & time &  0.73 &  0.28 &        6.69 \\
            & time-freq. &  0.69 &  0.31 &        7.26 \\
            \midrule
IG & frequency &  0.60 &  0.32 &        \textbf{6.97} \\
            & time &  0.34 &  0.35 &        7.14 \\
            & time-freq. &  0.67 &  0.31 &        8.52 \\
            \midrule
G$\times$I & frequency &  0.51 &  0.38 &        \textbf{7.03} \\
            & time &  0.32 &  0.37 &        7.19 \\
            & time-freq. &  0.58 &  0.33 &        8.iele58 \\
            \midrule
Sensitivity & frequency &  0.36 &  0.59 &        7.66 \\
            & time &  0.51 &  0.41 &        \textbf{6.13} \\
            & time-freq. &   0.36 &  0.59 &        9.96 \\
\bottomrule
\end{tabular}
\label{tab:audiomnist}
\end{table*}

\subsection{Use case I: Data representations for audio classifiers} \label{sec:use_case_1}
The best choice of data representation -- e.g. raw waveforms, spectrograms or spectral features -- is an important aspect of deep learning-based audio analysis \cite{dl_audio}. Previous work benchmarks different representations by measuring classification accuracy \cite{audio_representations, gupta_esc}. Novelly, DFT-LRP allows for a comparison of the underlying strategies of two audio classifiers trained in the frequency and time domain, which we leverage in this case study to demonstrate the utility of our approach. 

Here, we compare the 1d CNN sex audio classifier operating on the raw waveforms of the AudioMNIST dataset (\textit{time model}), to a model of the architecture, but trained on absolute values of the signal in frequency domain (\textit{frequency model}). The frequency model achieves an accuracy of 98\% on the sex classification task (the time model has an accuracy of 92\%).

\begin{figure}[t]
	\center
	\includegraphics[width=0.9\linewidth]{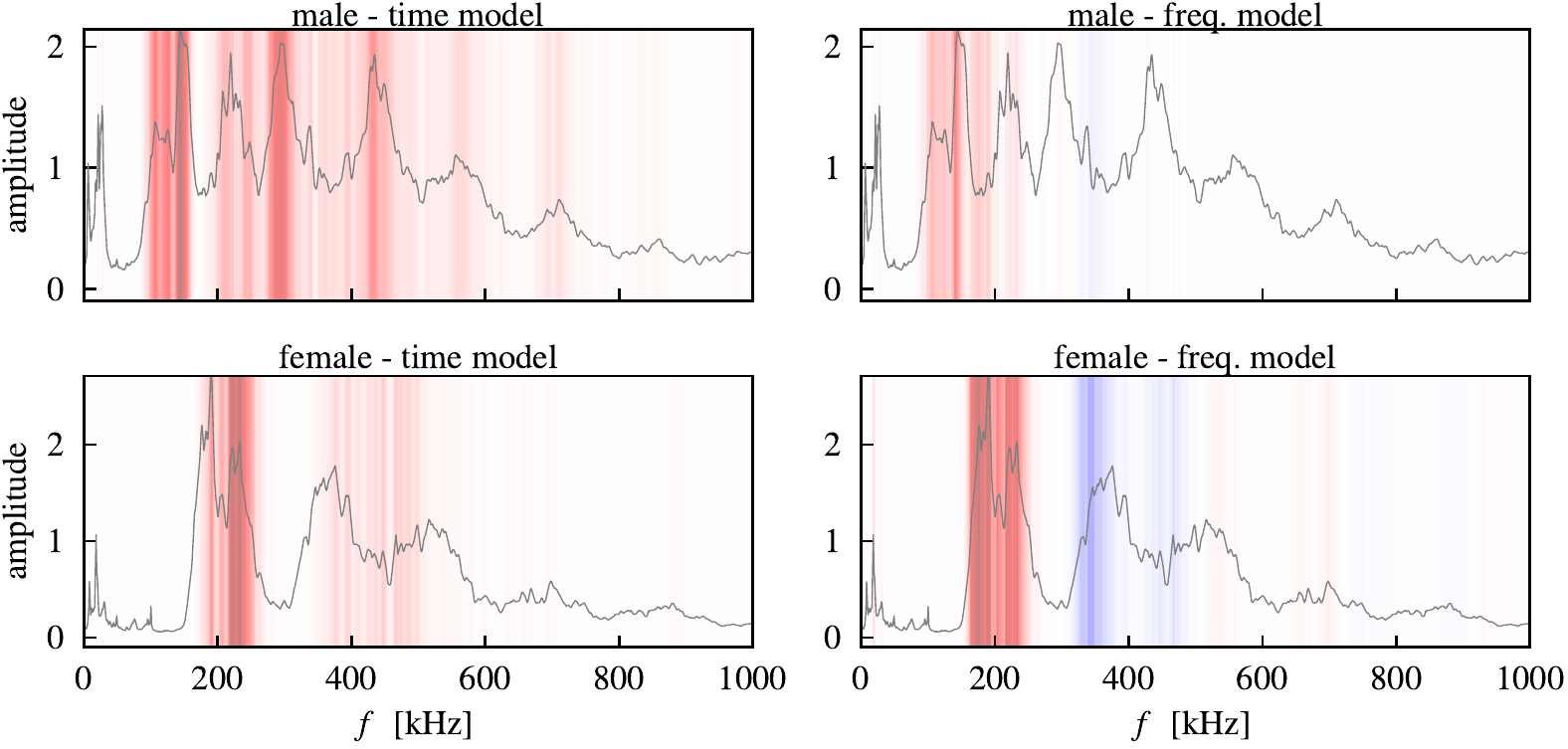}
	\caption{We evaluate LRP relevances for models trained on the AudioMNIST sex classification task. We show the mean spectrum and mean DFT-LRP relevances across the test dataset for male/female samples for the time and frequency model. The time model uses fundamental frequency and subsequent harmonics as features while the frequency model focuses only on the fundamental frequency. }
	\label{fig:t-f_audiomnist}
\end{figure}

To compare the classification strategies of the two models, we show the mean relevance in frequency domain for female and male samples for both models in \Cref{fig:t-f_audiomnist} across $3000$ test set samples (1500 female). The correlation between the relevances of the frequency and time model in the frequency domain is only 0.43 on average, already revealing that the two models have picked up different classification strategies. Before we look into these in more detail, we list the characteristics of female and male voices from the literature: The fundamental frequency of the male voice is between 85-155 Hz for males and 165-255 Hz for females \cite{fitch1970modal}, the subsequent harmonics are integer-multiples of this value.
To quantify the classification strategy of the two models, we list the frequency bands for which the mean relevance exceeds the 90\% percentile.
For male samples and the time model, the mean relevance exceeds the 90\% percentile for frequency intervals $99-156\;$Hz, $276-307\;$Hz, and $425-438\;$Hz.
For male samples and the frequency model, this is the case for 83-160 Hz, plus for a small number 24 frequencies in the intervals 293-300, 335-340 Hz, and 423-436 Hz. For the female samples, the analog threshold is exceeded between intervals 182-255 Hz and 392-487 Hz for the time model and between 18-20Hz (noise) and 157-253 Hz for the frequency model.
For comparison, the fundamental frequency of the male voice is between 85-155 Hz for males and 165-255 Hz for females \cite{fitch1970modal}, the subsequent harmonics are integer-multiples of this value.
In summary, the time model focuses on the fundamental frequency and the first two (one) subsequent harmonics of the male (female) voice, whereas the frequency model considers mostly the fundamental frequency as a relevant feature for male and female samples and interestingly also low frequencies corresponding to noise for the female samples.

\subsection{Use case II: DFT-LRP reveals Clever hans strategies in frequency domain} \label{sec:use_case_2}

\subsubsection{Artificial noise in audio data}
Noise is separated from the signal in frequency domain, but not in time domain.
We mimic a scenario which is realistic in various real world audio classification problems. We add noise to one class of the AudioMNIST digit classification task. Likely, noise as a spurious correlation also exists in real-world data due to class-dependent recording techniques or environment. A model that learns to separate classes by spurious correlations is deemed a Clever hans classifier \cite{cleverhans}. Here, we demonstrate, that Clever hans strategies leveraging noise can only be detected after propagating relevance from time to frequency domain.

\begin{figure}[t]
	\center
	\includegraphics[width=\linewidth]{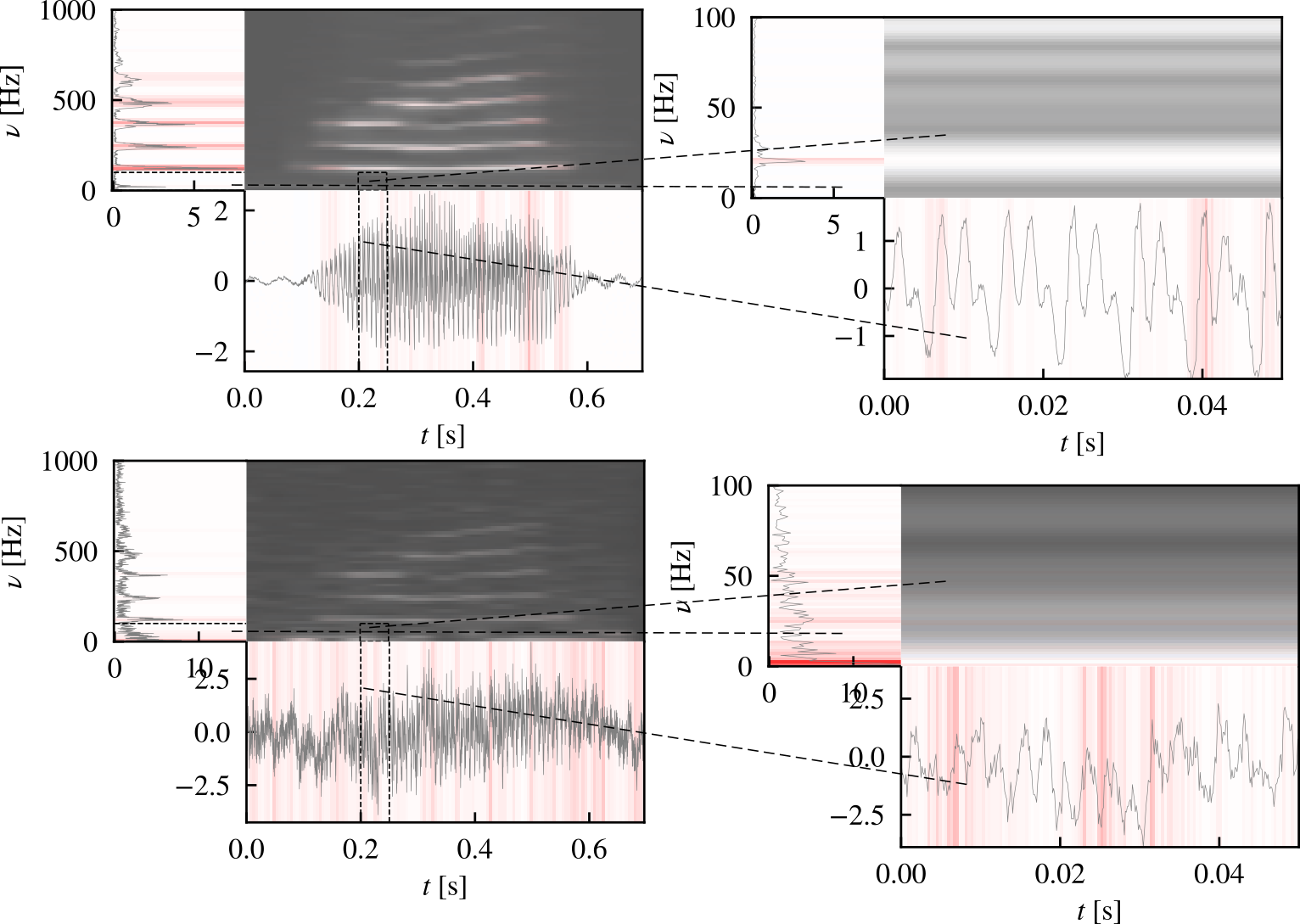}
	\caption{ We compare a model trained on data without additional noise (upper) to a model trained on data with Clever hans noise (lower), for a sample without and with added pink noise with strength $\sigma=0.8$. The left column shows the whole signal, while the right column shows a zoom on the part of the signal marked in the left column. All plots show the same sample (spoken zero). The noise was only added to 0s.}
	\label{fig:audiomnist_cleverhans}
\end{figure}

To this end, we compare a model trained on the original AudioMNIST digit classification data and a model trained on the modified data, where pink noise was added only to the spoken zeros. 
As in the previous section, both models are trained in the time domain and achieve an accuracy of about 94\%. We confirm that the model is using the Clever hans strategy, which we tried to induce by introducing the spurious correlation in the training data, by finding that it classifies 98\% of samples with added noise as zero, regardless of the actual digit spoken. Now, we try to infer this behavior from the explanations in \Cref{fig:audiomnist_cleverhans}, showing LRP relevances of the same sample with label zero for each classifier the in three domains. In the time domain, the only visible difference between the Clever hans and the regular classifier is that the beginning and end of the signal, where no digit is spoken, is relevant for the decision of the Clever hans but not for the regular model. Otherwise, relevance is spread rather uniformly over the signal in both cases. The difference between the classifiers only becomes perfectly clear in the time or time-frequency domain. The regular classifier focuses on the  fundamental frequencies and subsequent harmonics towards the beginning and end of the spoken digit, whereas the relevance of the Clever hans classifier is concentrated on frequencies between zero and 50 Hz, which correspond to noise.

\subsubsection{ECG classifier}
\begin{figure}[t]
	\center
	\includegraphics[width=0.49\linewidth]{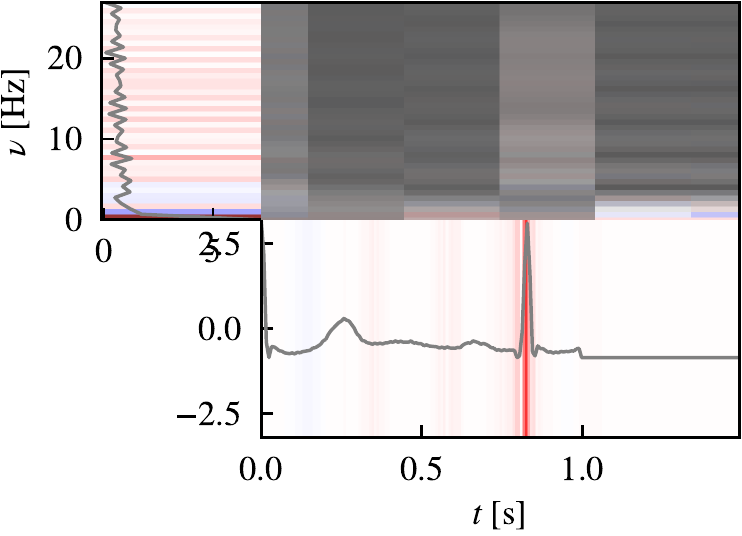}
    \includegraphics[width=0.49\linewidth]{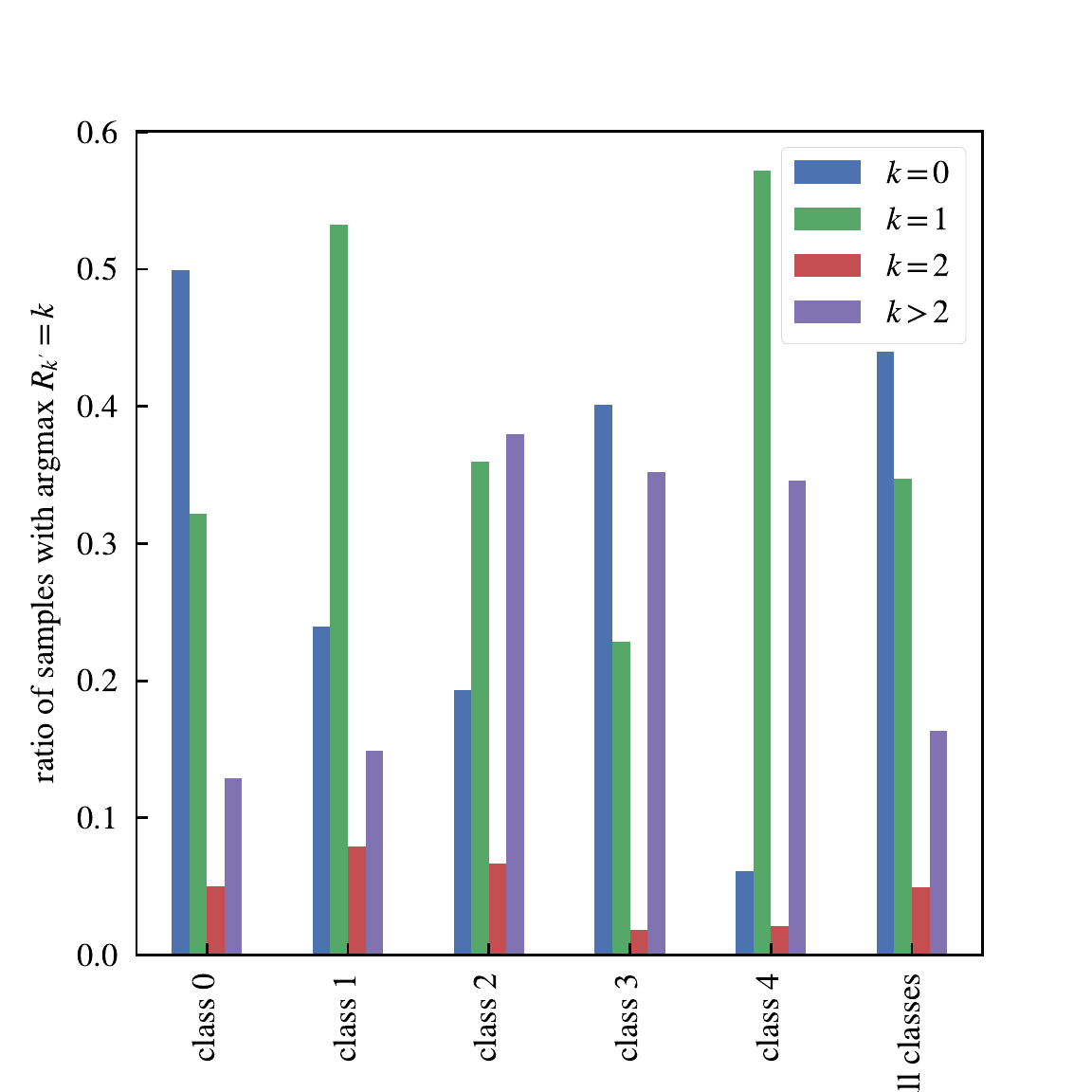}
	\caption{Left: LRP relevances for an ECG classifier trained on the MIT-BIH dataset in time, frequency and time-frequency domain for a correctly predicted random normal beat. Right: Ratio of samples, for which the maximum relevance $max_{k'} R_{k'}$ lies on the respective frequency component $k$. Among all classes, the classifier focuses on the mean of the signal ($k=0$) $44.3$\% of the samples. This value is even higher for class 0 (normal beats).}
	\label{fig:ecg_t-f-tf}
\end{figure}

We now demonstrate how DFT-LRP helps to discover Clever hans behaviour of the ECG classifier described in \Cref{sec:datasets}.
A trustworthy ECG model should base its classification strategy on similar signal characteristics like a cardiologist, such as the amplitude of the QRS complex and ST segment, the duration of segments or ratios of peaks. For instance, a normal heartbeat that originates at the atrium and traverses the normal conduction path is characterized by a sharp and narrow QRS complex with a broad peak at a frequency of 8\;Hz \cite{ecg_fourier}. We test the ECG classifier's strategy and start by showing ((ST)DFT)-LRP relevances in time, frequency and time-frequency domain for a normal beat in \Cref{fig:ecg_t-f-tf} (left). For visual clarity, we depict only frequency components up to $k=20$, where the most relevance is located. For this sample, relevance in time and time-frequency domain suggests that the model focuses on the QRS complex, in particular on frequencies around $k=10$. However, relevance in frequency domain shows that a large part of the total relevance is attributed to $k=0$ which corresponds to the mean of the signal. In \Cref{fig:ecg_t-f-tf} (right) we show the ratio of samples in the test set for which the maximum of $R_k$ lies on frequencies $k=0,1,2$ or $k>2$ for each class and observe that the model focuses on the mean of the signal for a majority of samples for all classes ($44.3$\%). This value is even higher for class 0 (normal beats). We conclude that the classifier learned to focus on the mean of the signal for these classes instead of characteristics considered by cardiologists, i.e. relies on a Clever hans strategy. Explanations for the model only reveal this behavior after transforming relevances from time to frequency domain via DFT-LRP.

\section{Conclusion}
We showed how to extend LRP for interpretable explanations for time series classifiers in frequency and time-frequency domain. After checking the validity of our approach in a ground-truth test bed and on real audio data, we demonstrated the benefits our methods bring in real-world settings, i.e. analysis of input representations and detection of Clever hans behaviour. 
We see applications of our method in domains where the time domain representation of the signal is particularly hard to intepret, like in audio, sensor data or electronic health records.
So far, we have focused on univariate time series. Although we can apply our method to multi-variate time series in a straightforward fashion by applying DFT-LRP to each channel separately, it is limited in the sense that it cannot reveal relevant interactions between channels, which could be an important aspect e.g. for ECG classifiers acting on multiple channels.
In future research, it would be interesting to test how other invertible transformations like PCA could serve as virtual inspection layers at the input or hidden feature layers of the model.
This could even be non-linear but approximately invertible transformations \cite{worrall2017interpretable, pmlr-v80-adel18a}, e.g. a learned interpretable representation via an auto encoder.

\section*{Acknowledgements}
This work was supported by the German Ministry for Education and Research (BMBF) through BIFOLD (refs. 01IS18025A and 01IS18037A), by the German Research Foundation under Grant DFG KI-FOR 5363, by the European Union’s Horizon 2020 research and innovation program under grants iToBoS (grant No. 965221) and TEMA (grant No. 101093003), and the state of Berlin within the innovation support program ProFIT (IBB) as grant BerDiBa II (grant no. 10185426).

\bibliographystyle{elsarticle-num}
\bibliography{bibfile}

\newpage
\appendix

\section{Inverse STDFT: WOLA vs. COLA} \label{sec:COLA}
If one does not re-scale the signal by the sum over all windows,
\begin{equation}
    \tilde x_n = \sum_m \idft(\{v_{m,k}\}) \, ,
\end{equation}
then,
\begin{equation*}
    \tilde x(n) = \sum_m x(n) \cdot w_m(n) \cdot w_m(n)\, ,
\end{equation*}
and we can read the condition for perfect reconstruction, i.e. $\tilde x(n)=x(n)$:
\begin{equation*} \label{eq:STDFT_condition}
    \sum_m w^2(m,n) = 1\,.
\end{equation*}
This is a stricter condition than for the WOLA technique.
It is full-filled for example by rectangular windows with any overlap or the half-cycle sine window with 50\% overlap. 

\end{document}